\documentclass[11pt]{article}

\usepackage[final]{acl}
\usepackage{times}
\usepackage{latexsym}

\usepackage[T1]{fontenc}

\usepackage[utf8]{inputenc}

\usepackage{microtype}

\usepackage{inconsolata}

\usepackage{graphicx}
\usepackage{times}
\usepackage{latexsym}
\usepackage{amsfonts}
\usepackage{enumitem}
\usepackage{amsmath}
\usepackage{booktabs}
\usepackage{multirow}
\usepackage{makecell}
\usepackage{tcolorbox}
\usepackage{longtable}
\usepackage{float}
\usepackage{tabularx}
\usepackage{algorithm}
\usepackage{algorithmic}
\tcbuselibrary{breakable} 
\usepackage[table]{xcolor}
\usepackage{bm}
\definecolor{qwenBlue}{HTML}{78A2E0}
\definecolor{codeblue}{RGB}{0,0,139}
\definecolor{codegreen}{RGB}{0,100,0}
\definecolor{codered}{RGB}{178,34,34}
\title{MARDoc: A Memory-Aware Refinement Agent Framework for Multimodal Long Document QA}

\author{
 \textbf{Kaifeng Chen\textsuperscript{1}},
 \textbf{Hongtao Liu\textsuperscript{2}},
 \textbf{Qiyao Peng\textsuperscript{1}}
 \thanks{Corresponding author.},
 \textbf{Jian Yang\textsuperscript{3}}\\
 \textbf{Yongqiang Liu\textsuperscript{1}},
 \textbf{Xiaochen Zhang\textsuperscript{4}},
 \textbf{Qing Yang\textsuperscript{2}},
\\
\\
 \textsuperscript{1}Tianjin University, Tianjin, China,
 \textsuperscript{2}Qifu Technology, Beijing, China\\
 \textsuperscript{3}Beihang University, Beijing, China, 
 \textsuperscript{4}Jiangnan University, Wuxi, China\\
 \small{
   \href{mailto:tiandapyy@tju.edu.cn}{tiandapyy@tju.edu.cn}, \href{mailto:htliu@tju.edu.cn}{htliu@tju.edu.cn}, \href{mailto:qypeng@tju.edu.cn}{qypeng@tju.edu.cn}, \href{mailto:jiaya@buaa.edu.cn}{jiaya@buaa.edu.cn}, 
   \href{mailto:lyq236@tju.edu.cn}{lyq236@tju.edu.cn}, 
 }\\
 \small{   
   \href{mailto:lyq236@tju.edu.cn}{lyq236@tju.edu.cn}, 
   \href{mailto:xiaochen_zhang@stu.jiangnan.edu.cn}{xiaochen\_zhang@stu.jiangnan.edu.cn}, 
   \href{mailto:workyang@gmail.com}{workyang@gmail.com}
   }
}

\begin{document}
\maketitle

\begin{abstract}
Iterative retrieval-reasoning agents have recently shown promise for multimodal long-document question answering. 
However, most existing systems maintain a single growing context that mixes retrieval traces, observations, and intermediate reasoning. As interactions accumulate, key evidence becomes scattered and diluted, making multi-hop reasoning noisy. 
We propose \textit{MARDoc}, a Memory-Aware Refinement Agent framework that decouples long-document QA into three specialized agents: an Explorer for multi-granularity multimodal retrieval, a Refiner for distilling interaction traces into structured evidence and reasoning memories, and a Reflector for checking evidence sufficiency and providing targeted feedback. 
Across iterations, the agents rely on a dynamically updated structured memory rather than a full accumulated interaction history.
This design reduces context noise while preserving answer-critical facts and their logical dependencies. 
Experiments on MMLongBench-Doc and DocBench show that \textit{MARDoc} achieves strong results, outperforming same-backbone baselines and demonstrating the effectiveness of structured memory for agentic document QA.
\end{abstract}
\section{Introduction}
Question answering over multimodal long documents is critical for many real-world applications, such as financial report analysis, scientific literature understanding, and enterprise document search. However, this task remains challenging for two reasons. First, information in real documents is jointly expressed through text, layout, tables, figures, and other visual elements, requiring models to align and reason across modalities. Second, supporting evidence is often sparsely distributed across many pages, making it difficult to locate, associate, and synthesize evidence over long contexts.

Existing research has mainly progressed along three lines. Early OCR-based document understanding models encode textual, layout, and visual signals in an end-to-end manner~\cite{xu2020layoutlm,xu2021layoutlmv2,huang2022layoutlmv3}. However, these methods are difficult to scale to very long documents due to context and input-length constraints. Retrieval-augmented methods retrieve relevant pages or document chunks before feeding them into multimodal large language models (MLLMs) for answer generation~\cite{guo2025rag,yu2024visrag}. While retrieval reduces the input length, evidence scattered across pages and modalities remains hard to associate, limiting multi-hop reasoning. More recently, agentic methods have introduced planning, tool invocation, and iterative retrieval into document RAG~\cite{han2025mdocagent,singh2025agentic,dong2026doc}. 
These methods improve the flexibility of long-document exploration, but we argue that two important limitations remain.

First, existing methods follow a ``monolithic context stream'' paradigm~\cite{sun2025docagent,zhang2026docdancer}: at each turn, the agent appends new retrieval results and intermediate reasoning to a single ever-growing context. As turns accumulate, irrelevant interaction history dilutes the context and compresses the model's effective reasoning space, yet no mechanism exists to distill key evidence into a compact, reusable representation.

Second, existing agentic methods often lack memory-grounded quality control over the reasoning process~\cite{zhu2025doclens,zhang2026docdancer}. As a result, it remains difficult to identify whether an error comes from missing evidence, misinterpreted visual content, or flawed cross-page reasoning. Errors introduced in early turns may therefore persist across iterations and compound hallucinations in multi-hop reasoning chains.

To address both issues, we propose \textit{MARDoc} (Memory-Aware Refinement Agent for Document QA), which replaces the monolithic context stream with an Explore–Refine–Reflect loop driven by a dynamically updated structured memory. In each iteration, the Explorer retrieves multi-granularity evidence from the document; the Refiner distills the interaction traces into a structured memory that retains answer-critical facts and their logical dependencies; and the Reflector examines this memory to determine whether the current evidence suffices and whether targeted feedback should guide the next round of exploration. Across iterations, \textit{MARDoc} carries forward the structured memory as the persistent state, while raw exploration traces are used only as transient inputs for memory refinement. This memory-mediated interface prevents the full interaction history from being repeatedly propagated, keeping the reasoning state compact while preserving key evidence and reasoning dependencies.

To evaluate our approach, we conduct comprehensive experiments on two challenging long-context multimodal document benchmarks: MMLongBench-Doc and DocBench, comparing against a wide range of baselines. Experimental results show that, even without task-specific training, our method achieves performance comparable to state-of-the-art models. Our main contributions are summarized as follows:
\begin{itemize}[itemsep=0.5pt, parsep=0pt]
\vspace{-2mm}
    \item
    We propose \textit{MARDoc}, which replaces the monolithic context stream with a dynamically updated structured memory. By decoupling information retrieval from evidence digestion, \textit{MARDoc} keeps the reasoning context compact across iterations, fundamentally alleviating the context dilution problem in existing iterative retrieval-reasoning methods.
    \item
    We introduce two complementary mechanisms: a distillation process that compacts interaction traces into structured evidence and reasoning memories, and an iterative reflective assessment that verifies evidence sufficiency and generates targeted feedback for subsequent exploration. 
    \item 
    We conduct experiments on two challenging multimodal long-document benchmarks. Under the same backbone, \textit{MARDoc} consistently outperforms existing baselines and achieves performance competitive with methods employing closed-source models, validating the effectiveness of structured memory for agentic document QA.
\end{itemize}

\section{Related Work}
\textbf{Multimodal Document QA.} Document QA has evolved from single-modal short-text tasks to complex long-document reasoning~\cite{ke2025large,mathew2021docvqa}, where evidence is often scattered across pages and modalities. Early methods relied on OCR pipelines to fuse text and visual features for MLLMs~\cite{peng2022ernie}. As MLLMs improved, some works bypassed the OCR stage and processed document images directly~\cite{zhang2023internlm,yan2026docseeker,duan2025docopilot}. However, these end-to-end approaches were strictly limited by context window sizes. To overcome this constraint, RAG-based methods retrieve only relevant multimodal chunks~\cite{cho2024m3docrag,yu2024visrag}, but they often ignored document structure and struggled with multi-step reasoning. More recently, a line of works has integrated agent technology into the RAG system~\cite{singh2025agentic,xu2025comprehensive}. These works utilize the knowledge integration and  tool invocation capabilities of agents to enhance the accuracy of the model's responses~\cite{han2025mdocagent,zhu2025doclens}. DocAgent~\cite{sun2025docagent} constructs the documents into a structured document tree and equips the agent with various tools to achieve efficient retrieval of relevant evidence. DocDancer~\cite{wu2025doc} further uses synthetic data to improve the agent's tool-use proficiency. However, due to the accumulation paradigm of the context, most methods are unable to effectively stimulate the multi-hop reasoning ability of the model in the long multimodal DocQA task.

\noindent\textbf{Agent Memory.} 
Memory mechanisms have become essential for enabling LLM-based agents to perform efficient long-range reasoning. Recent studies show that structured memory improves both reasoning performance and stability over extended interactions by mitigating information loss~\cite{mei2025survey,hu2025memory}. For example, Agentic Reasoning~\cite{wu2025agentic} and A-Mem~\cite{xu2025mem} model tool-calling trajectories as structured reasoning memory, while ReasoningBank~\cite{ouyang2025reasoningbank} extracts transferable tool-usage patterns from historical calls to improve tool-use accuracy and efficiency. These methods are designed for general-purpose agents and focus on compressing generic interaction history or accumulating cross-task experience. Inspired by these works, we introduce a structured memory tailored for document QA that iteratively distills answer-critical facts and their reasoning chains from interaction traces, directly addressing information fragmentation, weak logical connections, and context explosion in iterative document reasoning.
\section{Method}
\subsection{Overall Framework}
\begin{figure*}[t]
  \centering
   \includegraphics[width=\linewidth]{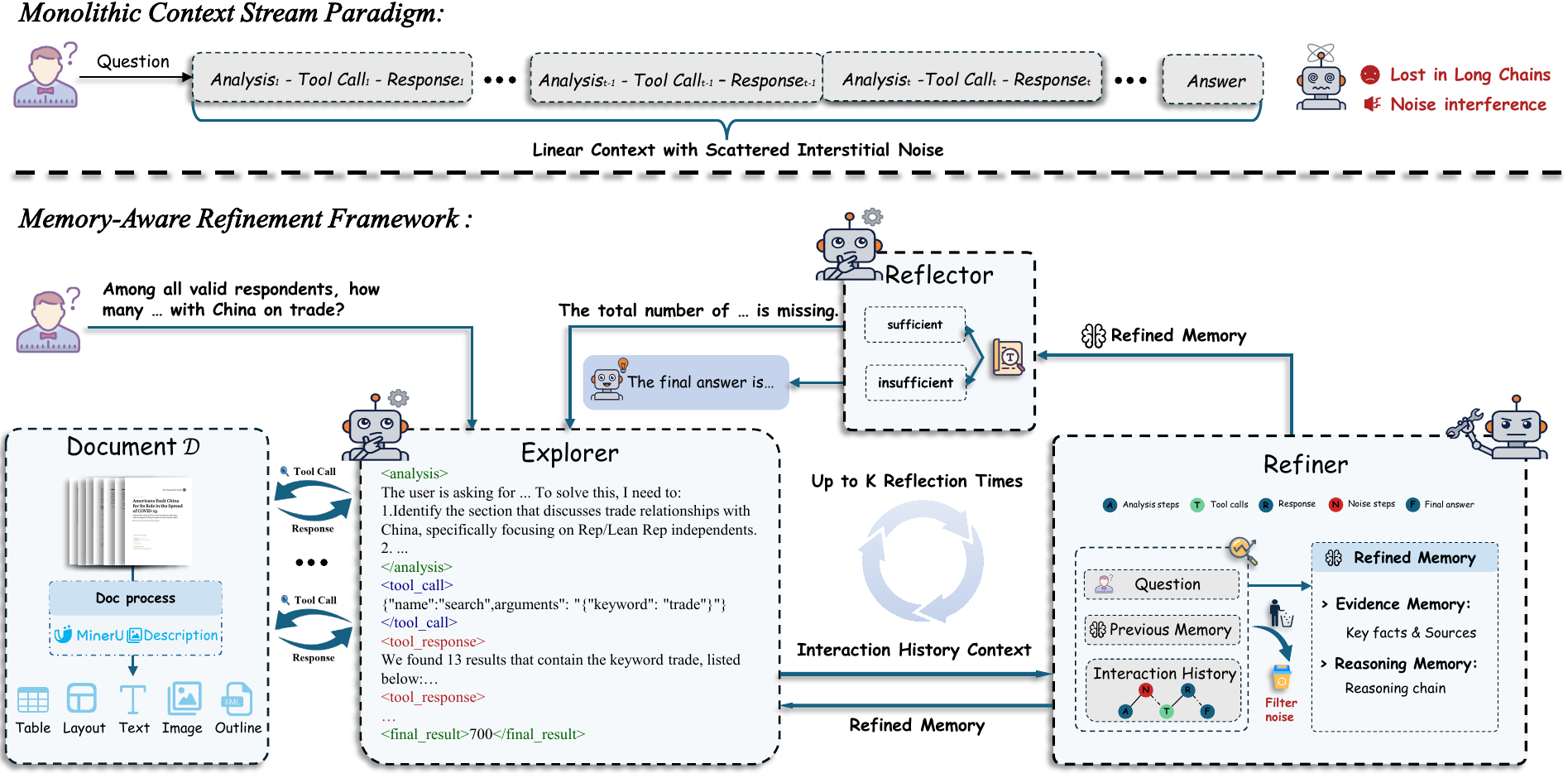}
   \caption{Overview of the \textit{MARDoc} framework.
\textbf{(Top)} The monolithic context stream paradigm appends all interactions to a single expanding context, causing key evidence to remain scattered without structured representation. \textbf{(Bottom)} The proposed \textit{MARDoc} framework decouples retrieval, refinement, and reflection into three specialized agents communicating through a dynamically updated report memory.We present a case study illustrating the whole process in Appendix~\ref{sec:caseStudy}.
}
   \label{fig:framework}
\end{figure*}

Given a user query $Q$ and a document $D=\{p_{1},p_{2},...,p_{n}\}$, where each $p_{i}$ denotes an individual page represented as an RGB image, the goal of document QA is to generate an accurate answer $A$ based on the evidence in the document $D$. As shown in Figure~\ref{fig:framework}, the overall reasoning process of \textit{MARDoc} follows an ``Explore-Refine-Reflect'' workflow. The three agents interact iteratively to progressively gather and refine the relevant evidence for the question. \textit{MARDoc} consists of four key components: (1) the outline construction module: generating a structured representation of the multimodal long document, providing the agent with rough information about the document; (2) the Explorer agent: using the outline and tools to retrieve multi-granularity relevant information and generate preliminary answers; (3) the Refiner agent: distilling structured evidence and reasoning memories from the interaction traces of the Explorer; (4) the Reflector agent: iteratively judging whether the current reasoning memories is correct and providing heuristic information to facilitate the Explorer to retrieve information better. In the following sections, we will discuss each component in detail.
\subsection{Document Processing} 
Prior XML-based document representations~\cite{sun2025docagent} organize parsed elements into hierarchical trees but may lose fine-grained layout and visual information. To ensure comprehensive coverage, we employ MinerU2.5~\cite{niu2025mineru2} for precise document parsing, which produces fine-grained semantic nodes preserving layout structure. We further utilize Qwen3-VL-235B-A22B-Instruct~\cite{bai2025qwen3} to generate concise
descriptions for visual elements, enriching the outline with retrievable visual content information. The outline serves as the unified index for all downstream retrieval tools.
\subsection{Explorer}
The Explorer iteratively retrieves evidence relevant to the user query from a structured document outline using customized tools. We adopt ReAct~\cite{yao2022react} as the agent framework. At each step, the agent generates a reasoning thought $\tau$ and an action $a$, forming a reasoning chain $\mathcal{T}$:
\begin{equation}
\mathcal{T}=(\tau_0,a_0,\varepsilon_0,\dots,\tau_i,a_i,\varepsilon_i,\dots,\tau_n,a_n),
\end{equation}
where $a_{n}$ denotes the terminal action in which the Explorer determines that sufficient information has been gathered and produces a candidate answer, $\varepsilon_i$ denotes the tool response at step $i$. At iteration $t$, the Explorer receives the query $Q$, the structured outline $\mathcal{O}$, the current memory $M_{t-1}$ from the Refiner, and the reflective instruction $R_{t-1}$ from the Reflector. The Explorer then integrates this information to conduct further retrieval, formulated as:
\begin{equation}
    \mathcal{T}_t=\mathcal{M}_{Exp}(Q,\mathcal{O},M_{t-1},R_{t-1})
\end{equation}
Here, $\mathcal{M}(\cdot)$ denotes the response generated by the large language model given a prompt. $\mathcal{T}_t$ represents the reasoning chain produced by the Explorer after the $t$-th iteration. 

To equip the Explorer with precise retrieval capabilities, we design a multi-granularity toolset that encompasses both textual and visual modalities. Detailed descriptions of the tools are provided in Appendix~\ref{imp_detail}.

\subsection{Refiner}
The Refiner is designed to simulate the reasoning mechanism of human experts when tackling complex problems. When solving multi‑step reasoning tasks, humans do not memorize every detail of a document; instead, they form a clear mental thread that records key information, core evidence, and the logical relationships among them. The Refiner explicitly models this cognitive process and distills context through explicit compacting, thereby replacing the traditional paradigm of monolithic context. At the end of each exploring iteration, the Refiner compacts the historical interaction traces and performs reasoning. In this process, the Refiner does not retain all interaction details; rather, it generates two structured memory components from the historical information. At iteration $t$, the Refiner takes three inputs: (1) the user query $Q$, (2) the previous memory $M_{t-1}$, and (3) the current reasoning chain $\mathcal{T}_{t}$ from the Explorer. By iteratively updating the memory, the Refiner efficiently integrates fragmented information, thereby enhancing the model’s reasoning capability:
\begin{equation}
    M_{E},M_{R} =\mathcal{M}_{Ref}(Q,\mathcal{T}_{t}, M_{t-1})
\end{equation}
The structured memory $M_{t}$ output by the Refiner consists of two distinct components: $M_{t}=(M_{E},M_{R})$. Here, $M_{E}$ denotes the Evidence Memory, and $M_{R}$ denotes the Reasoning Memory. These compacted Evidence Memory and Reasoning Memory replace the original monolithic context, enabling subsequent agents to review historical information in a compact manner while avoiding getting stuck in incorrect exploration paths.

\textbf{Evidence Memory.} This component serves as a high‑level log of the task, recording the most central and query‑relevant factual information distilled from the Explorer’s interaction traces. Each entry captures a key statement relevant to the question, accompanied by its source, while redundant information is removed. Multimodal information such as images and tables is uniformly condensed into simple evidence nodes, preventing the Reflector from being distracted by noise present in the raw context.

\textbf{Reasoning Memory.} This component extracts a clear reasoning path from the raw context and integrates it with the evidence nodes to form an evidence chain. Each node in the evidence chain is annotated with the evidence nodes on which it depends. This design decouples evidence from reasoning, allowing the Reflector to effectively trace whether an error originates from incorrect evidence nodes or flawed reasoning logic.

Once the current iteration is completed, the updated memory overwrites the old one; through this mechanism, the agent avoids continuous context accumulation and effectively filters out query‑irrelevant noise, so that the new context contains only the key information relevant to $Q$.

\subsection{Reflector}
The Reflector examines the compacted structure memory after each iteration and provides feedback to guide subsequent exploration. It is deployed immediately after the Refiner filters out irrelevant information, enabling timely identification of omissions or errors in the collected evidence. Given the current query $Q$ and the compacted memory $M_t$, the Reflector evaluates whether the gathered evidence is sufficient to produce a correct answer. If so, it outputs the final answer $A$:
\begin{equation}
    A=\mathcal{M}_{Ref}(M_{t},Q)
\end{equation}
Otherwise, if more information is needed or the current memory $M_t$ contains errors, it generates specific feedback to guide the Explorer toward more effective retrieval and prevent it from repeating past mistakes:
\begin{equation}
    R_t=\mathcal{M}_{Ref}(M_{t},Q)
\end{equation}
The instruction set $R_t$ summarizes key reasoning cues,such as ``\textit{the total number of something is missing}'', ``\textit{the user is asking about chatgpt4o not bert}''. These instructions can providing heuristic information to facilitate the Explorer to retrieve information better.
\section{Experiment}
\begin{table*}[t]
\centering
\small
\begin{tabular}{llccc}
\toprule

\multirow{2}{*}{\textbf{Method}} & \multirow{2}{*}{\textbf{Model}} & \multicolumn{2}{c}{\textbf{MMLongBench-Doc}}  & \multicolumn{1}{c}{\textbf{DocBench}}  \\
&&\textit{ACC}&\textit{F1} &\textit{LasJ} \\
\arrayrulecolor{black}
\hline
\rowcolor{qwenBlue!10} 
\multicolumn{5}{c}{\textit{MLLM-based Baseline}} \\
\arrayrulecolor{black!20}
\specialrule{\lightrulewidth}{0pt}{4pt}
DocSeeker~\cite{yan2026docseeker} & \texttt{Qwen-2.5-VL-7B-Instruct} & 40.1 & 38.4 &--\\
Docopilot~\cite{duan2025docopilot} & \texttt{InternVL2-8B} & 28.8 & 23.0 &--\\
\arrayrulecolor{black!20}
\midrule
\multirow{3}{*}{Vanilla VLM} 
& \texttt{GPT-4o} & 42.8 & 44.9& 63.1 \\
& \texttt{Qwen3-VL-30B-A3B-Instruct} & 45.9 & 45.6 & 63.2\\
& \texttt{Qwen3-VL-8B-Instruct} & 41.9 & 42.2 & 64.6\\
\arrayrulecolor{black}
\specialrule{\lightrulewidth}{1pt}{0pt}
\rowcolor{qwenBlue!10} 
\multicolumn{5}{c}{\textit{RAG-based Baseline}} \\
\arrayrulecolor{black!20}
\specialrule{\lightrulewidth}{0pt}{4pt}
SV-RAG ~\cite{chen2024sv} & \texttt{GPT-4o} & 23.0 & 24.2& --\\
VisRAG~\cite{yu2024visrag} & \texttt{GPT-4o} & 29.0 & 27.8  & -- \\
BOOKRAG~\cite{wang2025bookrag} & \texttt{Qwen2.5-VL-30B} & 43.8 & 44.9  & -- \\
MoLoRAG~\cite{wu2025molorag}  & \texttt{Qwen2.5-VL-7B} & 41.0 & -- & --\\
M3DocRAG~\cite{cho2024m3docrag} & \texttt{Qwen2-VL-7B} & 31.4 & 36.5 & -- \\
RAGAnything~\cite{guo2025rag} & \texttt{GPT-4o-mini} & 42.8 & --  & 63.4\\
\arrayrulecolor{black}
\specialrule{\lightrulewidth}{1pt}{0pt}
\rowcolor{qwenBlue!10} 
\multicolumn{5}{c}{\textit{Agent-based Baseline}} \\
\arrayrulecolor{black!20}
\specialrule{\lightrulewidth}{0pt}{4pt}
Doc-React~\cite{wu2025doc} & \texttt{GPT-4o} & 38.1 & 38.3 & -- \\
MDocAgent~\cite{han2025mdocagent} & \texttt{GPT-4o} &  42.0 & -- & -- \\
COA~\cite{zhang2024chain} & \texttt{Gemini 2.0 Flash} & 37.2 & 31.9  & -- \\
SLEUTH~\cite{liu2025resolving} & \texttt{Qwen3-VL-8B-Instruct} & 52.8 & --  & -- \\
\arrayrulecolor{black!20}
\midrule
\multirow{2}{*}{MACT~\cite{yu2025visual}}
& \texttt{MiMo-VL-7B} & 47.4 & --  & -- \\
& \texttt{Qwen2.5-VL-7B} & 43.7 & --  & -- \\
\arrayrulecolor{black!20}
\midrule
\multirow{4}{*}{DocAgent~\cite{sun2025docagent}} 
& \texttt{GPT-4o} & 51.8 & 49.1 & 79.9 \\
& \texttt{Claude 3.5 Sonnet} & \textbf{57.3} & \underline{54.1} & -- \\
& \texttt{Qwen3-VL-30B-A3B-Instruct} & 48.3 & 45.8 & 73.2 \\
& \texttt{Qwen3-VL-8B-Instruct} & 46.6 & 42.6 & 67.1 \\
\arrayrulecolor{black!20}
\midrule
\multirow{3}{*}{DocDancer~\cite{zhang2026docdancer}} 
& \texttt{GPT-4o} & 52.3 & 50.8 & 73.5\\
& \texttt{Qwen3-VL-30B-A3B-Thinking(ft)} & 54.4 & 53.9 & \underline{81.2}\\
& \texttt{Qwen3-VL-30B-A3B-Thinking} &39.2 & 36.4 & 74.1\\
\arrayrulecolor{black!20}
\specialrule{\lightrulewidth}{0pt}{2pt}
\multirow{2}{*}{\textbf{\textit{MARDoc}}} 
& \cellcolor{qwenBlue!60}\texttt{Qwen3-VL-30B-A3B-Instruct} & \cellcolor{qwenBlue!60}\underline{57.1} & \cellcolor{qwenBlue!60}\textbf{54.6} & \cellcolor{qwenBlue!60}\textbf{82.1} \\
& \cellcolor{qwenBlue!20}\texttt{Qwen3-VL-8B-Instruct} & \cellcolor{qwenBlue!20}52.7 & \cellcolor{qwenBlue!20}50.3 & \cellcolor{qwenBlue!20}72.3 \\
\arrayrulecolor{black}\midrule
Human Baseline  & --  &  65.8 & 66.0  & 81.2\\
\bottomrule
\end{tabular}
\caption{
Performance comparison across two multimodal long-document QA benchmarks. Performance results with best scores highlighted are \textbf{bolded} and second-best are \underline{underlined}. 
}
\label{tab:overall_result}
\end{table*}
\subsection{Experimental Setup}
We evaluate our framework with two backbone models: Qwen3-VL-30B-A3B-Instruct\footnote{In subsequent sections, we use Qwen3-30B to refer to Qwen3-VL-30B-A3B-Instruct.}~\cite{bai2025qwen3} and Qwen3-VL-8B-Instruct\footnote{We use Qwen3-8B to refer to Qwen3-VL-8B-Instruct.}~\cite{bai2025qwen3}. The initial Refiner memory $M_{0}$ and the initial Reflector instruction $R_{0}$ are empty string. Detailed implementation settings are provided in Appendix~\ref{imp_detail}.

\textbf{Benchmarks and Metrics.} We conduct experiments on two multimodal long-document QA benchmarks: MMLongBench-Doc~\cite{ma2024mmlongbench} and DocBench~\cite{zou2025docbench}. MMLongBench-Doc consists of 135 documents with an average length of 47.5 pages, covering seven domains with diverse layouts and multimodal content. It includes 1,082 questions derived from different sources and various locations, of which 34\% require cross-page reasoning. DocBench includes 229 real documents and 1,102 questions, spanning across five different domains and four major types of
questions.

For both benchmarks, we adopt their official evaluation protocols. For MMLongBench-Doc, we use GPT-4o to extract final answers and employ a rule-based method to compute Accuracy and F1 scores~\cite{ma2024mmlongbench}. For DocBench, we use the officially provided prompts to guide GPT-4o as the evaluator~\cite{zou2025docbench}.

\noindent\textbf{Baselines.} We compare our method against three categories of baselines:

(1) MLLM-based Methods: These approaches feed
all document pages images into MLLM and predict an answer directly, including Vanilla VLM based method, DocSeeker~\cite{yan2026docseeker} and Docopilot~\cite{duan2025docopilot}.

(2) RAG-based Methods: We compare against existing RAG frameworks for document QA, including SV-RAG~\cite{chen2024sv}, VisRAG~\cite{yu2024visrag}, BOOKRAG~\cite{wang2025bookrag}, MoLORAG~\cite{wu2025molorag}, M3DocRAG~\cite{cho2024m3docrag}, and RAGAnything~\cite{guo2025rag}.

(3) Agent-based Methods: We compare with existing agent frameworks for document QA, including Doc-React~\cite{wu2025doc}, MDocAgent~\cite{han2025mdocagent}, MACT~\cite{yu2025visual}, Chain-of-Agent~\cite{zhang2024chain}, SLEUTH~\cite{liu2025resolving}, DocAgent~\cite{sun2025docagent}, and DocDancer~\cite{zhang2026docdancer}.

Detailed descriptions of all baselines are provided in Appendix~\ref{sec:Baselines}.
\subsection{Overall Performance}
Table~\ref{tab:overall_result} reports the results on two multimodal long-document QA benchmarks. Based on the results, we make several important observations.

Agent-based methods consistently outperform both the MLLM-based baselines and the RAG-based baselines. MLLM-based methods answer long documents directly, while RAG-based methods incorporate retrieval modules. Both are unreliable in determining whether a question is answerable and struggle to integrate cross-page information logically. In contrast, agent-based methods overcome these shortcomings through multi-step planning, tool use, and explicit reasoning paths, leading to higher multi-hop accuracy.

However although existing agent-based methods have iterative advantage, most methods still suffer from the monolithic context paradigm. As interaction traces grow longer, key evidence becomes diluted by irrelevant observations. \textit{MARDoc} breaks the monolithic context paradigm through an iterative  ``Explore-Refine-Reflect'' cycle. It achieves an overall accuracy of 57.1\% with Qwen3-30B, matching DocAgent with Claude 3.5 Sonnet while using a much smaller backbone model. These findings indicate that \textit{MARDoc} successfully eliminates both misleading noise caused by extended historical contexts, suggesting that the proposed approach based on compacted structured evidence and reasoning memories enhances performance for DocQA. 
\subsection{Fine-Grained Performance Analysis}
\begin{figure}[t]
  \centering
   \includegraphics[width=\linewidth]{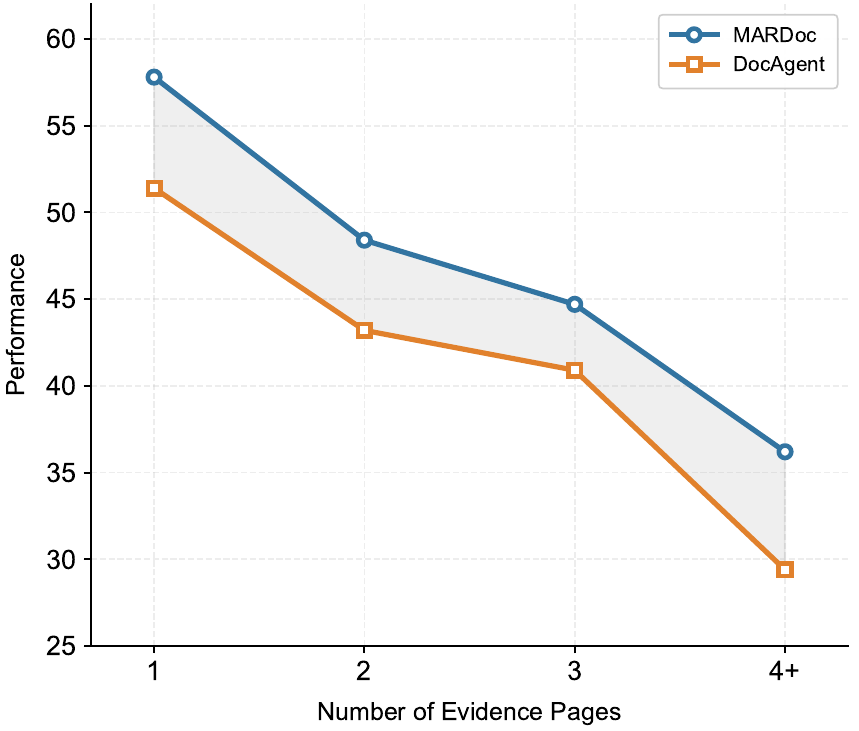}
   \caption{Accuracy on MMLongBench-Doc categorized
by the number of evidence pages.}
   \label{fig:evidence_page}
\end{figure}

We study the performance of models categorized by the number of pages that contain evidence which can be used to answer the question. The results are presented in Figure~\ref{fig:evidence_page}. As the number of evidence pages grows, both \textit{MARDoc} and DocAgent suffer from performance degradation. However, \textit{MARDoc} exhibits a notably smaller decline, suggesting that our structured memory enables the model to more accurately integrate information from diverse sources and maintain stronger multi-hop reasoning capability as evidence becomes scattered.
\subsection{Ablation Studies}
To evaluate the contribution of each component, we conduct ablation studies by progressively removing the Refiner and the Reflector. For a fair comparison, when both components are removed, we adopt a majority voting strategy that selects the most frequent answer across three inference runs as the final result. Table~\ref{ablation} shows the performance comparison across different document types.

The full \textit{MARDoc} achieves a substantial lead across all document domains, validating the effectiveness of the joint design of the Refiner and Reflector. Notably, retaining only the Reflector without the Refiner leads to drastic performance degradation, even falling below the variant where both components are removed. This suggests that without the structured memory compacted by the Refiner, the Reflector cannot effectively identify errors within the noisy, linearly accumulated context. In this case, excessive noise disrupts the reflection process itself, severely impairing overall performance.
\begin{table*}[t]
\centering
\begin{tabular}{l|ccccccc|ll}
\toprule
\multirow{2}{*}{\textbf{Method}}
& \multicolumn{7}{c|}{\textbf{Document Domain}} & \multirow{2}{*}{\textit{ACC}} & \multirow{2}{*}{\textit{F1}}\\
&\textit{Aca.}&\textit{Bro.}&\textit{Fin.}&\textit{Gui.}&\textit{Ind.}&\textit{Rep.}&\textit{Tut.}\\
\midrule
 \textit{MARDoc} & \textbf{54.9} & \textbf{49.3} & \textbf{68.8} &\textbf{52.2}&\textbf{66.1}&\textbf{57.4}&\textbf{55.5}&\textbf{57.1}&\textbf{54.6} \\ 
\quad \text{w/o} Refiner &42.3&44.2&40.0&45.2&58.0&46.6&49.5&45.9 &42.9\\
    \quad \text{w/o} Refiner \& Reflector   & 46.7&46.3&57.5&51.6&54.3& 50.5&47.2&50.2&47.7 \\
\bottomrule
\end{tabular}
\vspace{-0.2em}
\caption{Results of ablation studies on MMLongBench-Doc. We report the generalized accuracy of seven types of document domains, including Academia (Aca.), Brochure (Bro.), Finance (Fin.), Guidebook (Gui.), Industry (Ind.), Research (Rep.), Tutorial (Tut.). The best performance is highlighted in \textbf{bold}.}
\vspace{-0.8em}
\label{ablation}
\end{table*}

\subsection{Analysis of Refiner}
We conducted three ablation experiments to isolate the contribution of each memory component. In ablation (1), the Refiner outputs only the Evidence Memory $M_{E}$; in (2), only the Reasoning Memory $M_{R}$; and in (3), it outputs a plain-text compressed summary of the  interaction history, which serves as a baseline for unstructured memory. The results are reported in Table~\ref{wo:evidence-chain}.

The findings reveal complementary roles of the two memory components. Removing $M_{R}$ primarily degrades performance on multi-hop questions, indicating that $M_{R}$ maintains reasoning coherence when evidence must be synthesized across multiple pages. Removing $M_{E}$ substantially harms unanswerable question performance, suggesting that $M_{E}$ serves as a factual grounding mechanism that helps the Reflector distinguish insufficient evidence from contradictory premises, thereby reducing hallucinations. The plain-text compression baseline performs worst across all metrics, confirming that unstructured summarization retains noise and obscures logical structure. Together, $M_{E}$ ensures factual fidelity via verifiable source evidence, while $M_{R}$ ensures logical coherence by encoding reasoning dependencies; their combination mitigates hallucination and enables multi-hop reasoning.

\begin{table}[t]
\centering
\small
\begin{tabular}{l|ccc|cc}
\toprule
\multirow{2}{*}{\textbf{Method}}
& \multicolumn{3}{c|}{\textbf{Evidence Page}} & \multirow{2}{*}{\textit{ACC}} & \multirow{2}{*}{\textit{F1}}\\
&SIN&MUL&UNA\\
\midrule
 \textit{MARDoc} & \textbf{61.6} &\textbf{ 43.8} & \textbf{69.1} &\textbf{57.1}&\textbf{54.6} \\ 
\quad \text{w/o} $M_{R}$ &59.1&40.6&69.3&55.0&52.0\\
\quad \text{w/o} $M_{E}$ &59.4&42.2&64.7&54.9&52.3\\
\quad \text{w/o} $M_{E}$ \& $M_{R}$ &57.9&39.4&66.4&53.7&51.8\\
\bottomrule
\end{tabular}
\vspace{-0.2em}
\caption{Ablation on the structured memory of the Refiner: removing $M_{E}$, removing $M_{R}$, or replacing both with plain-text compression.}
\vspace{-0.8em}
\label{wo:evidence-chain}
\end{table}

\subsection{Analysis of Reflector}
\begin{figure}[t]
  \centering
   \includegraphics[width=\linewidth]{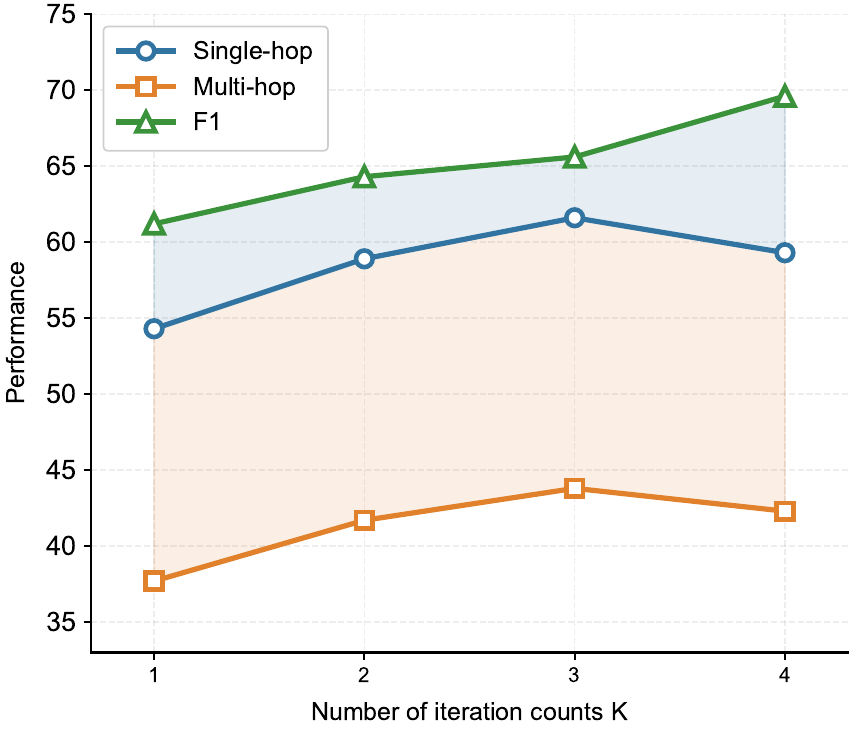}
   \caption{Performance variation with different maximum iteration counts $K$ of the Reflector agent.}
   \label{fig:num_k}
\end{figure}

To understand the effectiveness of the Reflector, we examine model performance under different maximum iteration counts $K$, as shown in Figure~\ref{fig:num_k}. We report results on three question types: single-hop, multi-hop, and unanswerable. From $K=1$ to $K=3$, the model achieves consistent improvements on both single-hop and multi-hop questions. However, when $K$ increases to 4, performance slightly declines. We attribute this to that further exploration rounds introduce redundant information that dilutes the previously structured memory, slightly degrading reasoning precision. Additionally, the F1 score gradually rises with more iterations, indicating that the Reflector partially mitigates hallucination issues.

We further analyze the impact of the instruction $R_{t}$ on performance. Table~\ref{wo_instruct} presents accuracy categorized by the number of evidence pages. In the ablation setting, the Reflector only evaluates the correctness of the reasoning memories without providing corrective instruction to the Explorer. Without the instruction, performance declines on both single-hop and multi-hop questions, while improving on the unanswerable type. We attribute this to the model's increased tendency to predict questions as unanswerable when explicit guidance is absent.

\begin{table}[t]
\centering
\small
\begin{tabular}{l|ccc|cc}
\toprule
\multirow{2}{*}{\textbf{Method}}
& \multicolumn{3}{c|}{\textbf{Evidence Page}} & \multirow{2}{*}{\textit{ACC}} & \multirow{2}{*}{\textit{F1}}\\
&SIN&MUL&UNA\\
\midrule
 \textit{MARDoc} & 61.6 & 43.8 & 69.1 &57.1&54.6 \\ 
\quad \text{w/o} instruct &56.7&40.8&69.5&53.9&50.8\\
    &\textcolor{blue}{\(\downarrow\)4.9}&\textcolor{blue}{\(\downarrow\)3.0}&\textcolor{blue}{\(\uparrow\)0.4}&\textcolor{blue}{\(\downarrow\)3.2}&\textcolor{blue}{\(\downarrow\)3.8}\\
\bottomrule
\end{tabular}
\vspace{-0.2em}
\caption{Results of ablation study on MMLongBench-Doc.We report different question type, inculding single-hop(SIN), multi-hop(MUL) and unanswerable(UNA).}
\vspace{-0.8em}
\label{wo_instruct}
\end{table}

\begin{figure}[t]
  \centering
   \includegraphics[width=\linewidth]{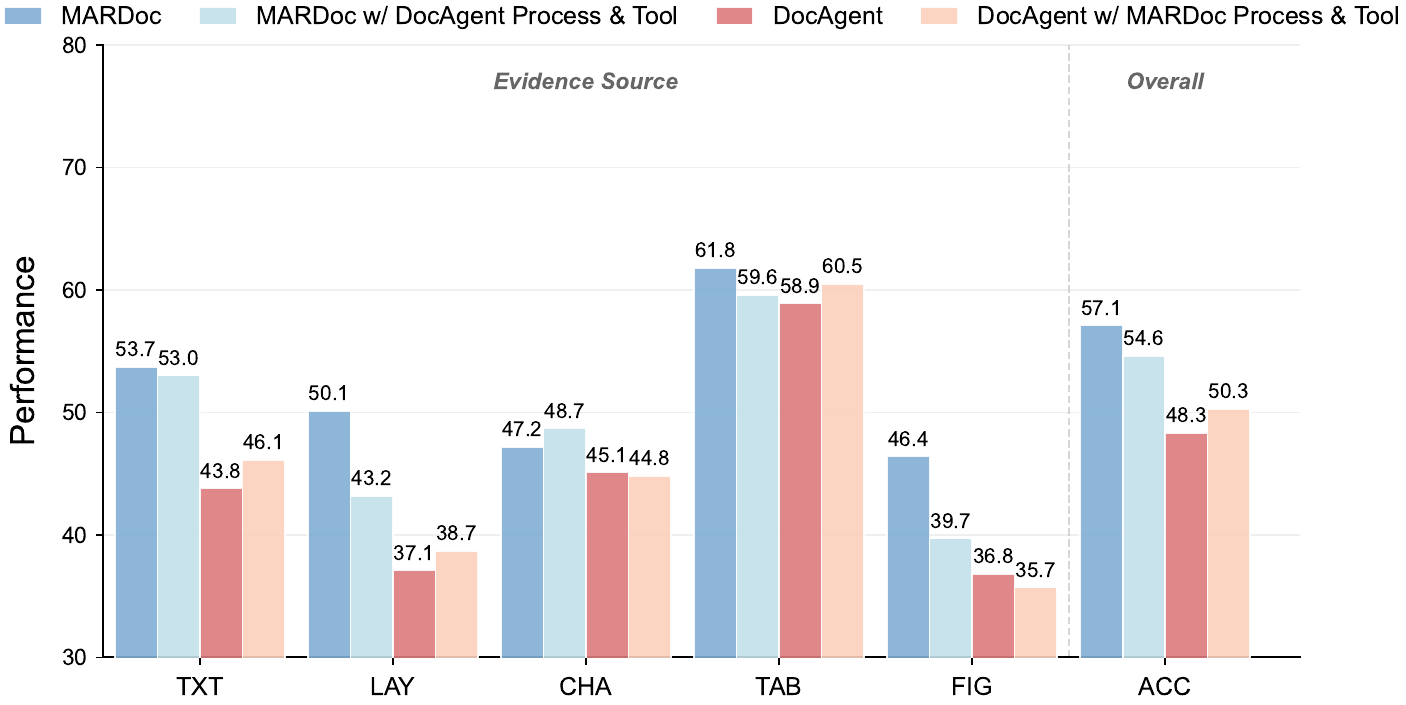}
   \caption{Ablation study on document process and tools. We report the generalized accuracy of five types of evidence sources, including Pure Text (TXT), Layout
(LAY), Chart (CHA), Table (TAB), and Figure (FIG).}
   \label{fig:ablation_xml}
\end{figure}
\subsection{Ablation study on document process and tools}
To validate the effectiveness of our outline construction and tool design, we conduct an ablation study comparing the document processing pipelines and toolsets of \textit{MARDoc} and DocAgent. For document parsing, DocAgent extracts PDF content using DocXChain~\cite{yao2023docxchain} and PyMuPDF, while \textit{MARDoc} employs MinerU2.5~\cite{niu2025mineru2} to construct outlines with finer-grained semantic nodes. For tool design, DocAgent provides five tools. We first adapted these tools to our new outline structure and then added two tools: get\_paragraph and search\_meta, to support multi-granularity retrieval.

As shown in Figure~\ref{fig:ablation_xml}, we evaluate performance by evidence source. Replacing DocAgent's outline with our outline and tools improves its performance across most evidence sources, with gains of 2.3\% for Pure Text, 1.6\% for Layout, and 1.6\% for Table. Although Chart and Figure performance decreases slightly by 1.4\%, overall accuracy improves by 2.0\%. Conversely, when \textit{MARDoc} uses DocAgent's outline, performance drops substantially, particularly in Layout and Figure. We attribute this to DocAgent's coarser document representation, which cannot effectively capture fine-grained layout and visual information. These results demonstrate that finer-grained document representations are essential for effective retrieval in agent-based document QA.
\subsection{Analysis of Computational Cost}
To evaluate practical efficiency and resource overhead, we conduct a cost analysis on 200 randomly sampled instances from MMLongBench-Doc using Qwen3-30B as the backbone model. Results are presented in Table~\ref{cost compare}. Compared to DocAgent, \textit{MARDoc} requires an additional 22.5k tokens and 15.9 seconds of latency per sample on average, while raising accuracy from 47.1\% to 54.9\%, a gain of 7.8\% points. This represents a favorable trade-off: the modest extra computation yields a substantial leap in accuracy on complex multi-page document understanding, consistent with the recent paradigm of test-time scaling and long chain-of-thought reasoning, where controlled additional investment in inference is rewarded with notably more reliable answers.

\begin{table}[t]
\centering
\setlength{\tabcolsep}{4pt}
\begin{tabular}{lccc}
\toprule
\textbf{Method} & \textbf{Avg.token(k)} & \textbf{Avg.times(s)} &\textbf{\textit{ACC}} \\ \midrule
 DocAgent & 100.3  & 49.4&47.1 \\ 
 \textit{MARDoc}  & 122.8  & 65.3&55.9\\
                  \bottomrule
\end{tabular}
\vspace{-0.2em}
\caption{Cost and efficiency comparison on MMLongDocBench.}
\vspace{-0.8em}
\label{cost compare}
\end{table}

\section{Conclusion}
We propose \textit{MARDoc}, a multi-agent framework for multimodal long-document question answering. \textit{MARDoc} decouples retrieval, evidence curation, and reflection into three specialized agents that communicate through a dynamically updated structured memory. The Refiner distills salient information from long interaction traces into an organized evidence structure, while the Reflector provides immediate instruction to mitigate noise accumulation during iterative retrieval. Experiments on MMLongBench-Doc and DocBench demonstrate that \textit{MARDoc} achieves strong and competitive performance, validating the effectiveness of memory-aware multi-agent collaboration for multimodal long-document QA.
\section*{Limitations}
Although MARDoc demonstrates promising results on multimodal
long-document QA, several limitations warrant discussion.
First, our agents are constructed entirely through prompt engineering without task-specific fine-tuning. Incorporating training-based optimization may further improve performance. Second, our evaluation is limited to the Qwen3-VL model family. Validating the generalizability of our framework across diverse model architectures remains an open direction. Third, the iterative Explore-Refine-Reflect loop involves multiple rounds of tool invocation and LLM inference, which increases system latency compared to single-pass methods. Exploring strategies such as early stopping could help mitigate this overhead.

\bibliography{custom}

@inproceedings{xu2020layoutlm,
  title={Layoutlm: Pre-training of text and layout for document image understanding},
  author={Xu, Yiheng and Li, Minghao and Cui, Lei and Huang, Shaohan and Wei, Furu and Zhou, Ming},
  booktitle={Proceedings of the 26th ACM SIGKDD international conference on knowledge discovery \& data mining},
  pages={1192--1200},
  year={2020}
}

@inproceedings{xu2021layoutlmv2,
  title={Layoutlmv2: Multi-modal pre-training for visually-rich document understanding},
  author={Xu, Yang and Xu, Yiheng and Lv, Tengchao and Cui, Lei and Wei, Furu and Wang, Guoxin and Lu, Yijuan and Florencio, Dinei and Zhang, Cha and Che, Wanxiang and others},
  booktitle={Proceedings of the 59th Annual Meeting of the Association for Computational Linguistics and the 11th International Joint Conference on Natural Language Processing (Volume 1: Long Papers)},
  pages={2579--2591},
  year={2021}
}

@inproceedings{huang2022layoutlmv3,
  title={Layoutlmv3: Pre-training for document ai with unified text and image masking},
  author={Huang, Yupan and Lv, Tengchao and Cui, Lei and Lu, Yutong and Wei, Furu},
  booktitle={Proceedings of the 30th ACM international conference on multimedia},
  pages={4083--4091},
  year={2022}
}

@article{guo2025rag,
  title={Rag-anything: All-in-one rag framework},
  author={Guo, Zirui and Ren, Xubin and Xu, Lingrui and Zhang, Jiahao and Huang, Chao},
  journal={arXiv preprint arXiv:2510.12323},
  year={2025}
}

@article{yu2024visrag,
  title={Visrag: Vision-based retrieval-augmented generation on multi-modality documents},
  author={Yu, Shi and Tang, Chaoyue and Xu, Bokai and Cui, Junbo and Ran, Junhao and Yan, Yukun and Liu, Zhenghao and Wang, Shuo and Han, Xu and Liu, Zhiyuan and others},
  journal={arXiv preprint arXiv:2410.10594},
  year={2024}
}

@article{cho2024m3docrag,
  title={M3docrag: Multi-modal retrieval is what you need for multi-page multi-document understanding},
  author={Cho, Jaemin and Mahata, Debanjan and Irsoy, Ozan and He, Yujie and Bansal, Mohit},
  journal={arXiv preprint arXiv:2411.04952},
  year={2024}
}

@inproceedings{sun2025docagent,
  title={Docagent: An agentic framework for multi-modal long-context document understanding},
  author={Sun, Li and He, Liu and Jia, Shuyue and He, Yangfan and You, Chenyu},
  booktitle={Proceedings of the 2025 Conference on Empirical Methods in Natural Language Processing},
  pages={17712--17727},
  year={2025}
}

@article{zhang2026docdancer,
  title={DocDancer: Towards Agentic Document-Grounded Information Seeking},
  author={Zhang, Qintong and Lv, Xinjie and Wu, Jialong and Li, Baixuan and Tao, Zhengwei and Yan, Guochen and Zhang, Huanyao and Wang, Bin and Xu, Jiahao and Mi, Haitao and others},
  journal={arXiv preprint arXiv:2601.05163},
  year={2026}
}

@inproceedings{wu2025doc,
  title={Doc-react: Multi-page heterogeneous document question-answering},
  author={Wu, Junda and Xia, Yu and Yu, Tong and Chen, Xiang and Harsha, Sai Sree and Maharaj, Akash V and Zhang, Ruiyi and Bursztyn, Victor and Kim, Sungchul and Rossi, Ryan A and others},
  booktitle={Proceedings of the 63rd Annual Meeting of the Association for Computational Linguistics (Volume 2: Short Papers)},
  pages={67--78},
  year={2025}
}

@article{ke2025large,
  title={Large language models in document intelligence: A comprehensive survey, recent advances, challenges, and future trends},
  author={Ke, Wenjun and Zheng, Yifan and Li, Yining and Xu, Hengyuan and Nie, Dong and Wang, Peng and He, Yao},
  journal={ACM Transactions on Information Systems},
  volume={44},
  number={1},
  pages={1--64},
  year={2025},
  publisher={ACM New York, NY}
}

@inproceedings{peng2022ernie,
  title={Ernie-layout: Layout knowledge enhanced pre-training for visually-rich document understanding},
  author={Peng, Qiming and Pan, Yinxu and Wang, Wenjin and Luo, Bin and Zhang, Zhenyu and Huang, Zhengjie and Cao, Yuhui and Yin, Weichong and Chen, Yongfeng and Zhang, Yin and others},
  booktitle={Findings of the Association for Computational Linguistics: EMNLP 2022},
  pages={3744--3756},
  year={2022}
}

@article{chen2024sv,
  title={SV-RAG: LoRA-contextualizing adaptation of MLLMs for long document understanding},
  author={Chen, Jian and Zhang, Ruiyi and Zhou, Yufan and Yu, Tong and Dernoncourt, Franck and Gu, Jiuxiang and Rossi, Ryan A and Chen, Changyou and Sun, Tong},
  journal={arXiv preprint arXiv:2411.01106},
  year={2024}
}

@article{zhang2023internlm,
  title={Internlm-xcomposer: A vision-language large model for advanced text-image comprehension and composition},
  author={Zhang, Pan and Dong, Xiaoyi and Wang, Bin and Cao, Yuhang and Xu, Chao and Ouyang, Linke and Zhao, Zhiyuan and Duan, Haodong and Zhang, Songyang and Ding, Shuangrui and others},
  journal={arXiv preprint arXiv:2309.15112},
  year={2023}
}

@article{mei2025survey,
  title={A survey of context engineering for large language models},
  author={Mei, Lingrui and Yao, Jiayu and Ge, Yuyao and Wang, Yiwei and Bi, Baolong and Cai, Yujun and Liu, Jiazhi and Li, Mingyu and Li, Zhong-Zhi and Zhang, Duzhen and others},
  journal={arXiv preprint arXiv:2507.13334},
  year={2025}
}

@article{hu2025memory,
  title={Memory in the age of ai agents},
  author={Hu, Yuyang and Liu, Shichun and Yue, Yanwei and Zhang, Guibin and Liu, Boyang and Zhu, Fangyi and Lin, Jiahang and Guo, Honglin and Dou, Shihan and Xi, Zhiheng and others},
  journal={arXiv preprint arXiv:2512.13564},
  year={2025}
}

@article{wu2025agentic,
  title={Agentic reasoning: Reasoning llms with tools for the deep research},
  author={Wu, Junde and Zhu, Jiayuan and Liu, Yuyuan},
  journal={arXiv preprint arXiv:2502.04644},
  volume={9},
  year={2025}
}

@article{xu2025mem,
  title={A-mem: Agentic memory for llm agents},
  author={Xu, Wujiang and Liang, Zujie and Mei, Kai and Gao, Hang and Tan, Juntao and Zhang, Yongfeng},
  journal={arXiv preprint arXiv:2502.12110},
  year={2025}
}

@article{ouyang2025reasoningbank,
  title={Reasoningbank: Scaling agent self-evolving with reasoning memory},
  author={Ouyang, Siru and Yan, Jun and Hsu, I and Chen, Yanfei and Jiang, Ke and Wang, Zifeng and Han, Rujun and Le, Long T and Daruki, Samira and Tang, Xiangru and others},
  journal={arXiv preprint arXiv:2509.25140},
  year={2025}
}

@inproceedings{yao2022react,
  title={React: Synergizing reasoning and acting in language models},
  author={Yao, Shunyu and Zhao, Jeffrey and Yu, Dian and Du, Nan and Shafran, Izhak and Narasimhan, Karthik R and Cao, Yuan},
  booktitle={The eleventh international conference on learning representations},
  year={2022}
}

@article{ma2024mmlongbench,
  title={Mmlongbench-doc: Benchmarking long-context document understanding with visualizations},
  author={Ma, Yubo and Zang, Yuhang and Chen, Liangyu and Chen, Meiqi and Jiao, Yizhu and Li, Xinze and Lu, Xinyuan and Liu, Ziyu and Ma, Yan and Dong, Xiaoyi and others},
  journal={Advances in Neural Information Processing Systems},
  volume={37},
  pages={95963--96010},
  year={2024}
}

@inproceedings{zou2025docbench,
  title={Docbench: A benchmark for evaluating llm-based document reading systems},
  author={Zou, Anni and Yu, Wenhao and Zhang, Hongming and Ma, Kaixin and Cai, Deng and Zhang, Zhuosheng and Zhao, Hai and Yu, Dong},
  booktitle={Proceedings of the 4th International Workshop on Knowledge-Augmented Methods for Natural Language Processing},
  pages={359--373},
  year={2025}
}

@article{wang2025bookrag,
  title={BookRAG: A Hierarchical Structure-aware Index-based Approach for Retrieval-Augmented Generation on Complex Documents},
  author={Wang, Shu and Zhou, Yingli and Fang, Yixiang},
  journal={arXiv preprint arXiv:2512.03413},
  year={2025}
}

@inproceedings{wu2025molorag,
  title={Molorag: Bootstrapping document understanding via multi-modal logic-aware retrieval},
  author={Wu, Xixi and Tan, Yanchao and Hou, Nan and Zhang, Ruiyang and Cheng, Hong},
  booktitle={Proceedings of the 2025 Conference on Empirical Methods in Natural Language Processing},
  pages={14035--14056},
  year={2025}
}

@article{han2025mdocagent,
  title={Mdocagent: A multi-modal multi-agent framework for document understanding},
  author={Han, Siwei and Xia, Peng and Zhang, Ruiyi and Sun, Tong and Li, Yun and Zhu, Hongtu and Yao, Huaxiu},
  journal={arXiv preprint arXiv:2503.13964},
  year={2025}
}

@article{yu2025visual,
  title={Visual document understanding and question answering: A multi-agent collaboration framework with test-time scaling},
  author={Yu, Xinlei and Chen, Zhangquan and Zhang, Yudong and Lu, Shilin and Shen, Ruolin and Zhang, Jiangning and Hu, Xiaobin and Fu, Yanwei and Yan, Shuicheng},
  journal={arXiv e-prints},
  pages={arXiv--2508},
  year={2025}
}

@article{bai2025qwen3,
  title={Qwen3-vl technical report},
  author={Bai, Shuai and Cai, Yuxuan and Chen, Ruizhe and Chen, Keqin and Chen, Xionghui and Cheng, Zesen and Deng, Lianghao and Ding, Wei and Gao, Chang and Ge, Chunjiang and others},
  journal={arXiv preprint arXiv:2511.21631},
  year={2025}
}

@article{hurst2024gpt,
  title={Gpt-4o system card},
  author={Hurst, Aaron and Lerer, Adam and Goucher, Adam P and Perelman, Adam and Ramesh, Aditya and Clark, Aidan and Ostrow, AJ and Welihinda, Akila and Hayes, Alan and Radford, Alec and others},
  journal={arXiv preprint arXiv:2410.21276},
  year={2024}
}

@article{liu2025resolving,
  title={Resolving evidence sparsity: Agentic context engineering for long-document understanding},
  author={Liu, Keliang and Chen, Zizhi and Li, Mingcheng and Tang, Jingqun and Yang, Dingkang and Zhang, Lihua},
  journal={arXiv preprint arXiv:2511.22850},
  year={2025}
}

@article{niu2025mineru2,
  title={Mineru2. 5: A decoupled vision-language model for efficient high-resolution document parsing},
  author={Niu, Junbo and Liu, Zheng and Gu, Zhuangcheng and Wang, Bin and Ouyang, Linke and Zhao, Zhiyuan and Chu, Tao and He, Tianyao and Wu, Fan and Zhang, Qintong and others},
  journal={arXiv preprint arXiv:2509.22186},
  year={2025}
}

@article{zhu2025doclens,
  title={Doclens: A tool-augmented multi-agent framework for long visual document understanding},
  author={Zhu, Dawei and Meng, Rui and Chen, Jiefeng and Li, Sujian and Pfister, Tomas and Yoon, Jinsung},
  journal={arXiv preprint arXiv:2511.11552},
  year={2025}
}

@article{zhang2024chain,
  title={Chain of agents: Large language models collaborating on long-context tasks},
  author={Zhang, Yusen and Sun, Ruoxi and Chen, Yanfei and Pfister, Tomas and Zhang, Rui and Ar{\i}k, Sercan {\"O}},
  journal={Advances in Neural Information Processing Systems},
  volume={37},
  pages={132208--132237},
  year={2024}
}

@inproceedings{dong2026doc,
  title={Doc-researcher: A unified system for multimodal document parsing and deep research},
  author={Dong, Kuicai and Huang, Shurui and Ye, Fangda and Han, Wei and Zhang, Zhi and Li, Dexun and Li, Wenjun and Yang, Qu and Wang, Gang and Wang, Yichao and others},
  booktitle={Proceedings of the ACM Web Conference 2026},
  pages={2349--2360},
  year={2026}
}

@article{xu2025comprehensive,
  title={A comprehensive survey of deep research: Systems, methodologies, and applications},
  author={Xu, Renjun and Peng, Jingwen},
  journal={arXiv preprint arXiv:2506.12594},
  year={2025}
}

@article{singh2025agentic,
  title={Agentic retrieval-augmented generation: A survey on agentic rag},
  author={Singh, Aditi and Ehtesham, Abul and Kumar, Saket and Khoei, Tala Talaei and Vasilakos, Athanasios V},
  journal={arXiv preprint arXiv:2501.09136},
  year={2025}
}

@inproceedings{mathew2021docvqa,
  title={Docvqa: A dataset for vqa on document images},
  author={Mathew, Minesh and Karatzas, Dimosthenis and Jawahar, CV},
  booktitle={Proceedings of the IEEE/CVF winter conference on applications of computer vision},
  pages={2200--2209},
  year={2021}
}

@article{yan2026docseeker,
  title={DocSeeker: Structured Visual Reasoning with Evidence Grounding for Long Document Understanding},
  author={Yan, Hao and Liu, Yuliang and Liu, Xingchen and Zhang, Yuyi and Liao, Minghui and Wu, Jihao and Chen, Wei and Bai, Xiang},
  journal={arXiv preprint arXiv:2604.12812},
  year={2026}
}

@inproceedings{duan2025docopilot,
  title={Docopilot: Improving multimodal models for document-level understanding},
  author={Duan, Yuchen and Chen, Zhe and Hu, Yusong and Wang, Weiyun and Ye, Shenglong and Shi, Botian and Lu, Lewei and Hou, Qibin and Lu, Tong and Li, Hongsheng and others},
  booktitle={Proceedings of the Computer Vision and Pattern Recognition Conference},
  pages={4026--4037},
  year={2025}
}

@article{yao2023docxchain,
  title={Docxchain: A powerful open-source toolchain for document parsing and beyond},
  author={Yao, Cong},
  journal={arXiv preprint arXiv:2310.12430},
  year={2023}
}
\clearpage
\newpage
\appendix
\section{Implementation Details}
    \label{imp_detail}
    \subsection{Tool description}
    \label{sub:tools}
    This section provides a detailed introduction to the tools offered to Explorer. We designe tools with different granularities and modalities to assist Explorer in effectively conducting document searches. The descriptions of these tools are shown in Table~\ref{tools}.
    
    \begin{table*}[t]
        \centering
        \small
        \begin{tabularx}{\textwidth}{llX}
            \toprule
            \textbf{Tool name} & \textbf{Parameter} & \textbf{Description}\\
            \midrule
            search & keyword & Find and extract all paragraphs and sections where the exact keyword appears  \\ 
            get\_section & section\_id &Get the full text content of a section in XML format  \\ 
            get\_paragraph & paragraph\_id & get the content of a specific paragraph by its ID  \\ 
            get\_page\_images & start\_page\_idx, end\_page\_idx & Extract full-page images from a specified range of PDF pages  \\ 
            get\_image &image\_id&Get the visual content of an image\\
            get\_table\_image & table\_id&Get the screenshot of a table\\
            search\_metadata&element\_types, keyword&Search for elements (text, titles, captions, images, tables) by keyword across all pages using PDF element metadata\\
            \bottomrule
        \end{tabularx}
        \vspace{-0.2em}
        \caption{Overview of the Tool Set for the Explorer. These tools empower the Explorer to perform coarse-to-fine evidence localization across textual, visual, and metadata modalities.}
        \vspace{-0.8em}
        \label{tools}
    \end{table*}

    \subsection{Details on Prompts for Agent}
    \label{Prompts}
    The prompts utilized for our three agent are presented in 
    Figure~\ref{fig:Explorer_prompt}, Figure~\ref{fig:Refiner_Prompt} and Figure~\ref{fig:Reflector_prompt}.
    \subsection{Inference Details}
    We deploy all backbone models using the vLLM inference framework to enable efficient large-scale multi-turn agent reasoning. We employ a temperature of 0.7, a top$_{p}$ value of 0.8. All experiments are conducted on four NVIDIA A800 GPUs (80GB). We set the maximum context length to 128K tokens for all methods unless otherwise specified.
    
    For the MARDoc framework, the maximum number of reflection iterations is set to $K=3$, and the maximum number of tool calls within each exploration stage is limited to 10 to prevent excessively long reasoning trajectories. All experiments are conducted in a training-free setting without any task-specific fine-tuning.

\section{Baselines}
\label{sec:Baselines}
We compare our method against a comprehensive set of baselines categorized into three groups:

\textbf{MLLM-based Baselines}:  We compare MARDoc
against two types of MLLM baselines for long-context document VQA.
\begin{itemize}
\vspace{-2mm}
    \item \textbf{Vanilla VLM.} This approach evaluates the vanilla long-context understanding capabilities of VLMs. We convert PDF pages into images at 144 DPI and directly feed them into the models without relying on external OCR parsing or retrieval modules. Following the setup of MMLongBench-Doc, we report the performance of GPT-4o~\cite{hurst2024gpt}, Qwen3-VL-30B-A3B and Qwen3-VL-8B~\cite{bai2025qwen3}. 
    \item \textbf{Processed VLM.} This approach also feeds all document pages as images, but applies additional processing on the VLM side. \textbf{DocSeeker}~\cite{yan2026docseeker} mitigates multi-page document VQA performance degradation from low SNR and sparse supervision using an ALR reasoning paradigm, two-stage training, and evidence-guided resolution allocation. \textbf{Docopilot}~\cite{duan2025docopilot} builds Doc750K, a multi-page dataset featuring cross-page dependencies and structured proxy tasks, and employs SFT to natively equip MLLMs for long-context document processing.
\end{itemize}

\textbf{RAG-based Baselines.} We compare MARDoc against diverse DocQA frameworks that employ various retrieval strategies, which can be broadly categorized into two paradigms:
\begin{itemize}
\vspace{-2mm}
    \item \textbf{Visual Retrieval.} \textbf{SV-RAG}~\cite{chen2024sv} utilizes the hidden embeddings of LLMs for question-based retrieval, enhanced by LoRA adapters. \textbf{VisRAG}~\cite{yu2024visrag} leverages visual embeddings of document images to retrieve relevant pages, maximizing the retention and utilization of raw visual information. \textbf{M3DocRAG}~\cite{cho2024m3docrag} employs ColPali for vision-based page retrieval and utilizes Qwen2-VL to conduct question answering.
    \item \textbf{Structure aware Retrieval.} \textbf{MoLoRAG}~\cite{wu2025molorag} constructs a page relation graph to perform VLM-driven image retrieval based on both semantic and logical relevance. \textbf{RAGAnything}~\cite{guo2025rag} constructs a cross-modal knowledge graph with non-textual units as nodes and a textual knowledge graph via entity and relation extraction, integrating graph structures with vector semantic matching for hybrid retrieval. \textbf{BOOKRAG}~\cite{wang2025bookrag} extracts a hierarchical tree from documents to serve as a table of contents, and builds a knowledge graph to capture complex inter-entity relations. By mapping entities to tree nodes, it designs an intelligent agent to execute customized textual retrieval processes.
\end{itemize}

\textbf{Agent-based Baselines.} We compare MARDoc against agent frameworks specifically designed for DocQA, which can be categorized into multi-agent collaboration and iterative retrieval paradigms.
\begin{itemize}
\vspace{-2mm}
    \item \textbf{Multi-Agent Collaboration.} \textbf{MDocAgent}~\cite{han2025mdocagent} employs a multi-agent architecture that separately processes different modalities to answer queries and integrates the results into a final answer. \textbf{MACT}~\cite{yu2025visual} decomposes the task into four specialized agents, incorporating a hybrid reward mechanism and a two-stage training strategy.
    \textbf{Chain-of-Agent}~\cite{zhang2024chain}is an agent-based framework in which multiple worker agents handle text segments one after another, and a manager agent then combines their outputs to produce the final response.SLEUTH~\cite{liu2025resolving} retrieves top‑K pages, extracts clues, filters irrelevant pages, and adaptively reasons over a distilled multimodal context to answer long‑document questions.
    \item \textbf{Iterative Retrieval.} \textbf{Doc-React}~\cite{wu2025doc} utilizes an agent with iterative retrieval decisions to balance information gain and uncertainty reduction. \textbf{DocAgent}~\cite{sun2025docagent} leverages a structured document tree outline and builds interactive multimodal retrieval tools, introducing a reviewer agent for cross-checking answers and maintaining a task-agnostic memory bank. \textbf{DocDancer}~\cite{zhang2026docdancer} designs lightweight dual tools based on the ReAct framework and utilizes a data synthesis pipeline to generate high-quality training data, resulting in an end-to-end trained open-source document agent.
\end{itemize}

\section{Scaling to Stronger Backbones}
\begin{table}[t]
\centering
\small
\setlength{\tabcolsep}{3.5pt}
\begin{tabular}{lccc}
\toprule
 \multirow{2}{*}{\textbf{Model}} & \multicolumn{2}{c}{\textbf{MMLongBench-Doc}}  & \multicolumn{1}{c}{\textbf{DocBench}}  \\
&\textit{ACC}&\textit{F1} &\textit{LasJ} \\
\arrayrulecolor{black}
\midrule
Qwen3-VL-30B-A3B&57.1&54.6&82.1\\
Qwen3-VL-32B&\textbf{59.9}&\textbf{57.5}&\textbf{83.2}\\
\bottomrule
\end{tabular}
\caption{Results of MARDoc with Qwen3-VL-32B-Instruct on MMLongBench-Doc and DocBench.
}
\label{tab:scaling}
\end{table}

To investigate the scalability of MARDoc with more capable backbone models, we additionally evaluate our framework using Qwen3-VL-32B-Instruct as backbone. As shown in
Table~\ref{tab:scaling}, MARDoc achieves an accuracy of 59.9\% on MMLongBench-Doc,
further improving over the Qwen3-VL-30B-A3B-Instruct result reported
in the main experiments. These results surpasses DocAgent with
Claude 3.5 Sonnet, demonstrating that our structured memory mechanism can effectively leverage the enhanced reasoning capabilities of stronger models. 

\section{Case Study}
\label{sec:caseStudy}
This section illustrates the collaborative problem-solving process of the three agents of MARDoc through a concrete example. In the first iteration, misled by noise, the Explorer yields an incorrect answer due to insufficient information retrieval. Subsequently, the Refiner extracts highly relevant evidence from the linearly stacked context to construct an evidence-grounded reasoning chain. Based on this, the Reflector analyzes the current state to identify the missing information required to answer the question. In the second iteration, leveraging the feedback from both the Refiner and Reflector, the Explorer successfully retrieves the missing information and derives the correct answer. Finally, the Refiner and Reflector jointly verify the correctness of this answer before outputting the final result.
\onecolumn
\definecolor{PromptFrame}{HTML}{516C94}  
\definecolor{PromptBack}{HTML}{E6EAEF}   
\begin{tcolorbox}[
  colback=PromptBack,
  colframe=PromptFrame,
  title=\textsc{Case Study of MARDoc},
  fontupper=\scriptsize,
  boxrule=0.5pt,
  breakable, 
]
{
\textbf{Question} \\
\quad Among all valid respondents,  how many rep/lean rep independents (the exact number, not the ratio) favor building a strong relationship with China on trade? Please give your answer to the nearest hundred, such as 100, 200 or 300....\\
\textbf{Answer}\\
300
\begin{tcolorbox}[
    colback=orange!90!black!80!,  %
    colframe=orange!90!black!80!, %
    boxrule=0.5pt,                %
    center,                       %
    fontupper=\footnotesize\bfseries, %
    halign=center,                %
    valign=center,                %
    colupper=white!90!,               %
    height=5mm,                   %
    left=2pt, right=2pt
]
Navigator
\end{tcolorbox}
    \begin{center}
        \textbf{\textit{{\small step 1}}}
    \end{center}
    
    \textcolor{codegreen}{\texttt{<analysis>}} \\
    \quad The user is asking for ...  To solve this, I need to:
    
    1. Identify the section that discusses trade relationships with China, specifically focusing on Rep/Lean Rep independents.\\
    2. Find the relevant data, which is likely in a table or chart showing opinions on trade with China.\\
    ...\\
    \textcolor{codegreen}{\texttt{</analysis>}} \\
    \textcolor{codeblue}{\texttt{<tool\_call>}} \\
    \quad\texttt{\{"name":"search",arguments": "\{"keyword": "trade"\}"\}}\\
    \textcolor{codeblue}{\texttt{</tool\_call>}} \\
    \textcolor{codered}{\texttt{<tool\_response>}} \\
    \quad We found 13 results that contain the keyword trade, listed below:\\
    \quad <Item type="Paragraph" section\_id="12" page\_idx="9">\\
    \quad ... including the loss of U.S. jobs to China (by 24 percentage points) and the U.S. trade deficit with China (19 points) ...\\
    \quad </Item>\\
    \quad \quad ...\\
    \quad <Section section\_id="47" page\_idx="27">\\
    \quad \quad ...\\ 
    \quad \quad <Image image\_id="25" page\_idx="27" page\_nums="26" description="presents survey findings on whether the U.S. public prefers to "get tougher" or "build a strong relationship" with China on economic and trade issues"/>\\
    \quad \quad <Paragraph page\_idx="28" paragraph\_id="210">... more say they were ultimately bad for the U.S. (44\% ) than good (30\% )  ...</Paragraph>\\
    \quad </Section>\\
    \quad </Item>\\
    \textcolor{codered}{\texttt{</tool\_response>}}
    \begin{center}
        \textbf{\textit{{\small step 2}}}
    \end{center}
    \textcolor{codegreen}{\texttt{<analysis>}} \\
    \quad The search results indicate that the relevant information about Rep/Lean Rep independents favoring building a strong relationship with China on trade is likely in Section 47. ... The chart in Image 25 on page 27 is likely to contain the detailed data needed to answer the question ...\\
    \textcolor{codegreen}{\texttt{</analysis>}} \\
    \textcolor{codeblue}{\texttt{<tool\_call>}} \\
    \quad \texttt{\{"name":"get\_section",arguments": "\{"section\_id": "47"\}"\}}\\
    \textcolor{codeblue}{\texttt{</tool\_call>}} \\
    \textcolor{codered}{\texttt{<tool\_response>}} \\
    Here is the full text content of Section 47:\\
    \quad <Section section\_id="47" start\_page\_idx="27" end\_page\_idx="28">\\
    \quad ...\\
    \quad \quad <Paragraph page\_idx="28" paragraph\_id="210">...When asked about the effects of increased tariffs on goods from foreign countries, including from China, more say they were ultimately bad for the U.S. (44\% ) than good (30\% ) , while about a quarter of Americans think the tariff's had no discernable effect on the U.S. ...</Paragraph>\\
    \quad </Section>\\
    \textcolor{codered}{\texttt{</tool\_response>}}
    \begin{center}
        \textbf{\textit{{\small step 3}}}
    \end{center}
    \textcolor{codegreen}{\texttt{<analysis>}} \\
    \quad The full content of Section 47 confirms that the relevant data is in the chart (Image 25) on page 27. ...\\
    \textcolor{codegreen}{\texttt{</analysis>}} \\
    \textcolor{codeblue}{\texttt{<tool\_call>}} \\
    \quad \texttt{\{"name":"get\_page\_images",arguments": "\{"start\_page\_idx": "47","end\_page\_idx": "47"\}"\}}\\
    \textcolor{codeblue}{\texttt{</tool\_call>}} \\
    \textcolor{codered}{\texttt{<tool\_response>}} \\
    \quad Here are the page images for page 27 to page 27:\\
    \begin{center}
        {\includegraphics[width=0.7\linewidth]{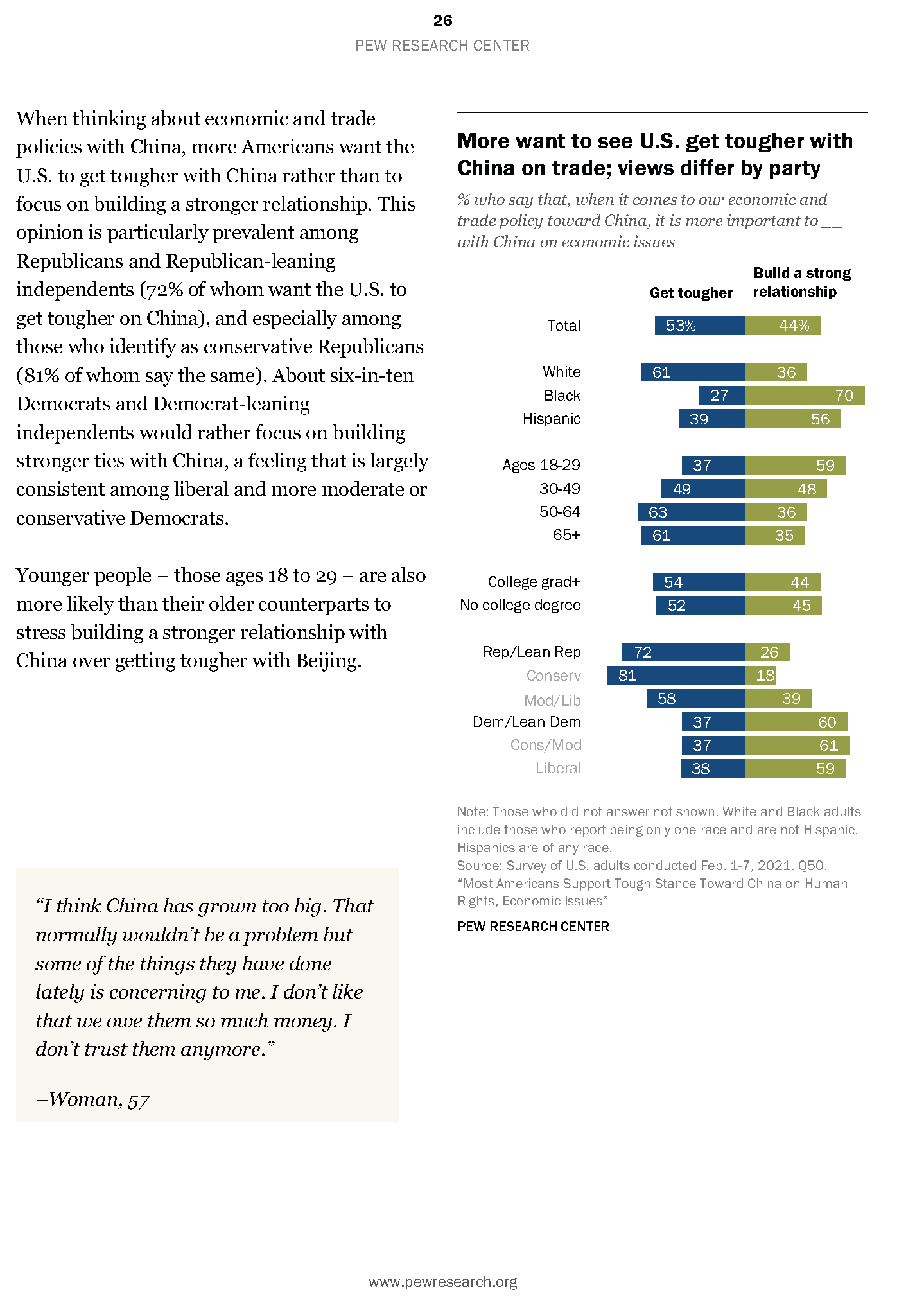}\par}     
    \end{center}
    \textcolor{codered}{\texttt{</tool\_response>}}
    \begin{center}
        \textbf{\textit{{\small step 4}}}
    \end{center}
    \textcolor{codegreen}{\texttt{<analysis>}} \\
    \quad The image of page 27 contains the chart and the relevant data. ... For Rep/Lean Rep, the percentage who favor building a strong relationship with China on trade is \textcolor{codegreen}{26\%}. To find the exact number of Rep/Lean Rep independents who favor building a strong relationship with China on trade, I need to know the total number of valid respondents. The document states that the survey included \textcolor{codered}{2,596} U.S. adults. Now, I will calculate the number of Rep/Lean Rep independents who favor building a strong relationship with China on trade by multiplying the total number of respondents by the percentage (26\%).Number of Rep/Lean Rep independents = 2,596 * 0.26 = 674.96.Rounding to the nearest hundred, the number is 700. I will now provide the final answer.\\
    \quad <final\_result>700</final\_result>\\
    \textcolor{codegreen}{\texttt{</analysis>}} 
\begin{tcolorbox}[
    colback=orange!90!black!80!,  %
    colframe=orange!90!black!80!, %
    boxrule=0.5pt,                %
    center,                       %
    fontupper=\footnotesize\bfseries, %
    halign=center,                %
    valign=center,                %
    colupper=white!90!,               %
    height=5mm,                   %
    left=2pt, right=2pt
]
Refiner
\end{tcolorbox}
Compact Evidence:\\
    \quad - The chart on page 27 shows the percentage of different political groups who believe it is more important to get tougher with China on economic issues versus building a strong relationship. (Source: Image on page 27)\\
    \quad - For Rep/Lean Rep, the percentage who favor building a strong relationship with China on trade is 26\%. (Source: Image on page 27)\\
    \quad - The survey included 2,596 U.S. adults. (Source: Text  on page 9)\\
    \quad ...\\
    \quad Logical Reasoning steps:\\
    \quad Step 1: Based on the chart on page 27, the percentage of Rep/Lean Rep independents who favor building a strong relationship with China on trade is 26\%.\\
    \quad Step 2: Based on the text on page 9, the total number of valid respondents in the survey is 2,596.\\
    \quad Step 3: To find the exact number of Rep/Lean Rep independents who favor building a strong relationship with China on trade, multiply the total number of respondents by the percentage (26\%).\\
    \quad Step 4: Calculate 2,596 * 0.26 = 674.96.\\
    \quad Step 5: Round 674.96 to the nearest hundred, which is 700.
\begin{tcolorbox}[
    colback=orange!90!black!80!,  %
    colframe=orange!90!black!80!, %
    boxrule=0.5pt,                %
    center,                       %
    fontupper=\footnotesize\bfseries, %
    halign=center,                %
    valign=center,                %
    colupper=white!90!,               %
    height=5mm,                   %
    left=2pt, right=2pt
]
Reflector
\end{tcolorbox}
\quad We already knows that for Rep/Lean Rep, the percentage who favor building a strong relationship with China on trade is 26\%. However,2,596 is the total number of valid respondents. The total number of Rep/Lean Rep individuals surveyed is missing.
\begin{tcolorbox}[
    colback=orange!90!black!80!,  %
    colframe=orange!90!black!80!, %
    boxrule=0.5pt,                %
    center,                       %
    fontupper=\footnotesize\bfseries, %
    halign=center,                %
    valign=center,                %
    colupper=white!90!,               %
    height=5mm,                   %
    left=2pt, right=2pt
]
Navigator
\end{tcolorbox}
    \begin{center}
        \textbf{\textit{{\small step 1}}}
    \end{center}
    \textcolor{codegreen}{\texttt{<analysis>}} \\
    \quad The question asks for... The key information provided indicates that:\\
    \quad 1. The survey included 2,596 U.S. adults.\\
    \quad 2. Image\_id=25 shows that 26\% of Rep/Lean Rep individuals favor building a strong relationship with China on trade.\\
    \quad 3. The total percentage for Rep/Lean Rep individuals in the chart is 98\% (72\% + 26\%), suggesting the percentages are based on the total number of Rep/Lean Rep individuals.\\
    \quad 4. The information does not provide the total number of Rep/Lean Rep individuals surveyed.\\
    \quad To answer the question, I need to find the total number of Rep/Lean Rep individuals surveyed. This information is likely in the section discussing the survey methodology or the breakdown of respondents by political affiliation. I will start by searching for relevant keywords to locate this information. \\
    \textcolor{codegreen}{\texttt{</analysis>}} \\
    \textcolor{codeblue}{\texttt{<tool\_call>}} \\
    \quad \texttt{\{"name":"search",arguments": "\{"keyword": "Rep/Lean Rep"\}"\}}\\
    \textcolor{codeblue}{\texttt{</tool\_call>}} \\
    \textcolor{codered}{\texttt{<tool\_response>}} \\
    \quad We found 4 results that contain the keyword Rep/Lean Rep, listed below:\\
    \quad <Search\_Result>\\
    \quad \quad ...\\
    \quad \quad <Item type="Table" section\_id="66" page\_idx="39" table\_id="5">\\
    \quad\quad Group Unweighted sample size Weighted \%  Plus or minus ...\\
    \quad\quad </Item>\\
    \quad </Search\_Result>\\
    \textcolor{codered}{\texttt{</tool\_response>}}
    \begin{center}
        \textbf{\textit{{\small step 2}}}
    \end{center}
    \textcolor{codegreen}{\texttt{<analysis>}} \\
    \quad The search results show that the total number of Rep/Lean Rep individuals surveyed is 1,106. This information is found in the table in section\_id="66" on page 39. ...\\
    \textcolor{codegreen}{\texttt{</analysis>}} \\
    \textcolor{codeblue}{\texttt{<tool\_call>}} \\
    \quad \texttt{\{"name":"get\_table\_image",arguments": "\{"table\_id": "5"\}"\}}\\
    \textcolor{codeblue}{\texttt{</tool\_call>}} \\
    \textcolor{codered}{\texttt{<tool\_response>}} \\
    \quad Here is the image content for table\_id 5:\\
    \begin{center}
        {\includegraphics[width=1\linewidth]{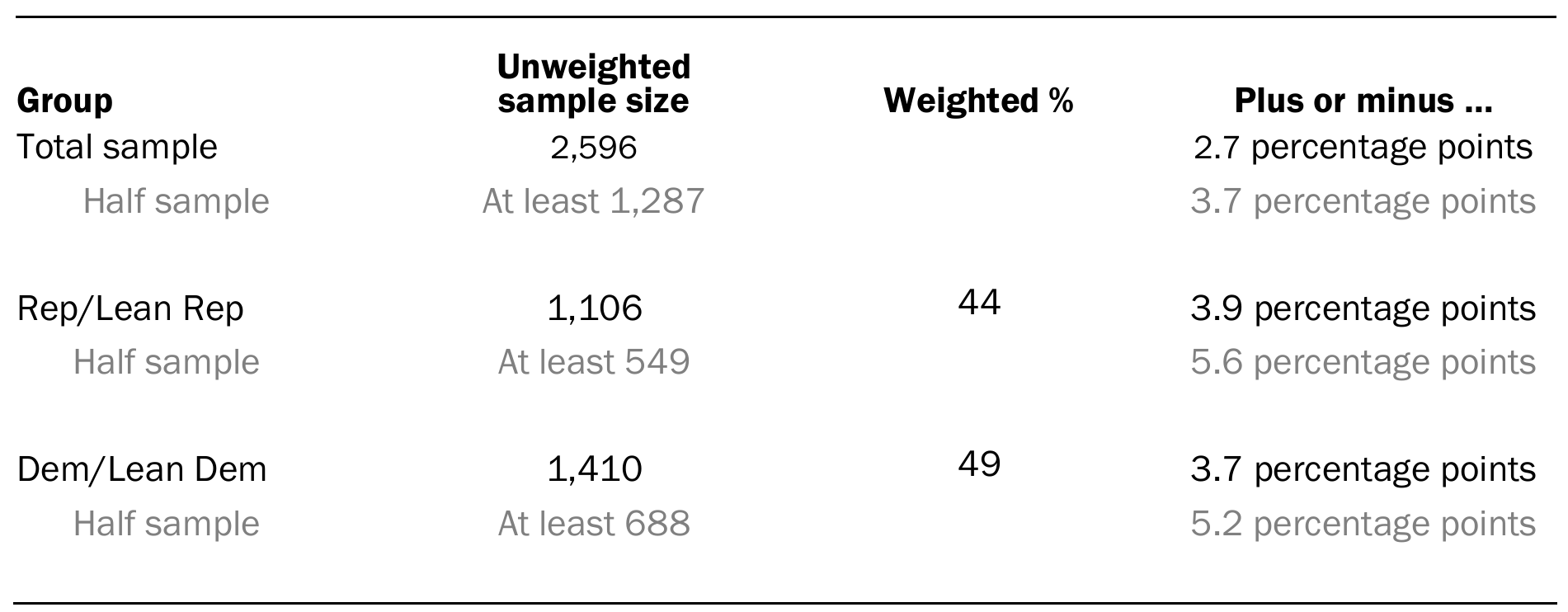}\par}
    \end{center}
    \textcolor{codered}{\texttt{</tool\_response>}}
    \begin{center}
        \textbf{\textit{{\small step 3}}}
    \end{center}
    \textcolor{codegreen}{\texttt{<analysis>}} \\
    The chart in image\_id="25" shows that 26\% of Rep/Lean Rep individuals favor building a strong relationship with China on trade. The table in section\_id="66" confirms that the total number of Rep/Lean Rep individuals surveyed is 1,106. Now I can calculate 26\% of 1,106 to find the exact number of individuals who favor building a strong relationship with China on trade. 26\% of 1,106 is 287.56, which rounds to 300 when rounded to the nearest hundred.\\
    <final\_result>\\
    300\\
    </final\_result>\\
    \textcolor{codegreen}{\texttt{</analysis>}}
\begin{tcolorbox}[
    colback=orange!90!black!80!,  %
    colframe=orange!90!black!80!, %
    boxrule=0.5pt,                %
    center,                       %
    fontupper=\footnotesize\bfseries, %
    halign=center,                %
    valign=center,                %
    colupper=white!90!,               %
    height=5mm,                   %
    left=2pt, right=2pt
]
Refiner
\end{tcolorbox}
Compact Evidence:\\
    - The chart in image\_id="25" shows that 26\% of Rep/Lean Rep individuals favor building a strong relationship with China on trade. (Source: Image 25)\\
    - The table in section\_id="66" shows that the unweighted sample size for Rep/Lean Rep individuals is 1,106. (Section 66)\\
    ...\\
Logical Reasoning steps:\\
    Step 1: The chart in image\_id="25" shows that 26\% of Rep/Lean Rep individuals favor building a strong relationship with China on trade\\
    Step 2: The table in section\_id="66" shows that the unweighted sample size for Rep/Lean Rep individuals is 1,106\\
    Step 3: The percentages in the chart are based on the total number of Rep/Lean Rep individuals, so the 26\% figure applies to the 1,106 individuals\\
    Step 4: To find the exact number, calculate 26\% of 1,106, which equals 287.56 \\
    Step 5: Rounding 287.56 to the nearest hundred gives 300 \\
\begin{tcolorbox}[
    colback=orange!90!black!80!,  %
    colframe=orange!90!black!80!, %
    boxrule=0.5pt,                %
    center,                       %
    fontupper=\footnotesize\bfseries, %
    halign=center,                %
    valign=center,                %
    colupper=white!90!,               %
    height=5mm,                   %
    left=2pt, right=2pt
]
Reflector
\end{tcolorbox}
    Based on the information in the memory, I will directly arrive at the final answer to the question.\\
    <final\_result>\\
    300\\
    </final\_result>\\
}
\end{tcolorbox}
\onecolumn
    \begin{figure*}[!htp]
        \centering
        
    \definecolor{PromptFrame}{HTML}{516C94}  
    \definecolor{PromptBack}{HTML}{E6EAEF}   
    
    \begin{tcolorbox}[
        breakable,
        title=Prompt for Explorer,
        colback=PromptBack,  
        colframe=PromptFrame, 
        fonttitle=\bfseries,
    ]
    \textbf{System Prompt}\\
    You are an expert AI assistant specializing in long document understanding. Given a XML format of a multimodal long document and a user question and a list of key information about the question, your task is using the given tools to systematically locate the indices of XML document that might contain information useful for answering the user’s question. \\
    Your core capabilities:
    \begin{itemize}[label=-,leftmargin=2em, nosep]
        \item Searching and locating relevant passages using provide tools.
        \item Understanding document structure through section hierarchies.
    \end{itemize}
    \textbf{User Prompt}\\
    I've uploaded a document, and below is the outline in XML format:\\
    \{document\_outline\}\\
    In the outline,page\_idx means the physics page and the page\_num means the printed page. Can you provide the relevant context about the following question based on the content of the document?\\
    \{question\}\\
    Here is the key information about the question:\\
    \{memory\}\\
    Here is the explorer instruction about the question:\\
    \{instruction\}\\
    Before solving the problem, you need to first formulate a plan to guide the subsequent tools calls.\\
    Important guidelines:
    \begin{itemize}[label= -,leftmargin=2em, nosep]
    \item Your primary responsibility is to provide the relevant information related to the issue obtained from the XML outline to the other agents. The other agents cannot directly view the XML outline; they can only see the output of your tool. Therefore, even if you obtain the relevant information solely from the XML outline, you still need to extract the corresponding context through the tool invocation.
    \item Before calling tools, please first deeply understand the user's question and break down the key points of the question.
    \item Before each step, wrap your thought process in <analysis></analysis> tags. This will help ensure a thorough and accurate analysis of the document and question.
    \item Even if you have already obtained some key information related to the question from the XML outline, you still need to use a tool to obtain the context of that key information to further confirm it.
    \end{itemize}
    \end{tcolorbox}
        \caption{Prompt for \textbf{Explorer}.}
        \label{fig:Explorer_prompt}
    \end{figure*}

    \begin{figure*}[!htp] 
        \centering
        
    \definecolor{PromptFrame}{HTML}{516C94}  
    \definecolor{PromptBack}{HTML}{E6EAEF}   
    
        \begin{tcolorbox}[
            title= Prompt for Refiner, 
            fontupper=\small,
            colback=PromptBack,    
            colframe=PromptFrame,  
            coltitle=white,
            label=fig:reflector_prompt
        ]
            \textbf{Sysyem Prompt}\\
            You are an expert in extracting key information, specializing in handling the reasoning steps in the field of information retrieval. Your task is to extract key information directly relevant to the user's question from the reasoning steps output by the document retrieval expert.\\
            \medskip
            \textbf{User Prompt}\\
            Please accurately and concisely extract key information and then conduct step-by-step reasoning to form a clear reasoning chain based on the previous memory and new relevant information.Update the memory with new information that helps to answer the problem, while retaining all relevant details from the previous memory\\
            \textbf{Input Format}\\
                Question: The problem posed by the user that needs to be solved.\\
                Previous memory: The memory of previous reasoning steps.\\
                Reasoning steps: The Reasoning steps in the previous round.\\
            \textbf{Important guidelines:}\\
            1. Deep Thinking:please first deeply understand the user's question.\\
            2. Comprehensive Perception and Information Extraction:
                \begin{itemize}[label= -,leftmargin=2em, nosep]
                    \item Image: Identify the text, objects, scenes, and states in the image and convert them into textual descriptions.
                    \item Chart: Extract the chart title, axis labels, legend, data points, trends, extreme values, proportions, etc.
                    \item Table: Extract key rows, columns, and cell data, and pay attention to the correspondence between the headers and the data.
                    \item Visual Relationships: Pay attention to the spatial relationships (such as above, below, left, right, inclusion), color distinctions, arrow directions, etc. of the implicit logic among elements
                \end{itemize}
            3. Evidence-Driven: Make decisions based on reasoning steps  deep thinking
                \begin{itemize}[label= -,leftmargin=2em, nosep]
                    \item Each piece of information should indicate from which part of the provided information it originates..
                    \item Each key point should be on a separate line, maintaining complete meaning and without adding any additional explanations or comments.
                \end{itemize}
            \medskip
            \textbf{Strictly follow these steps:}\\
            1.Extraction: Scan [Current Reasoning steps] and [Previous Memory]. \\
            2.Discard any irrelevant noise, redundant steps, or off-topic details. Extract only the core evidence nodes (facts, numbers, intermediate conclusions) that are logically connected to Q.\\
            3.Reasoning Reconstruction: Rebuild a highly condensed reasoning chain based only on the extracted evidence nodes. This chain must reflect the logical progression toward answering Q.\\
            4.Overwrite: Generate the new memory $M_t$. The new memory must completely replace the old memory to ensure timeliness and prevent context bloating. The new memory must be strictly limited to the essential extracted information and the reconstructed chain.\\
            \textbf{Input:}\\
            Current Question:\{question\}\\
            previous memory:\{memory\}\\
            the Reasoning steps are list below:\{Reasoning\_steps\}\\
            \textbf{Output Format:}\\
            Compact Evidence:\\
            - [Information Point 1]\\
            - [Information Point 2]\\
            Logical Reasoning steps:\\
            Step 1: [Based on information A and B, infer C]\\
            Step 2: [Based on C and information D, infer E]\\
            ...
            
        \end{tcolorbox}
        \caption{Prompt for \textbf{Refiner}.}
        \label{fig:Refiner_Prompt}
    \end{figure*}
    
    \begin{figure*}[!htp] 
        \centering
    \definecolor{PromptFrame1}{HTML}{516C94}  
    \definecolor{PromptBack1}{HTML}{E6EAEF}   
    
        \begin{tcolorbox}[
            title= Prompt for Reflector, 
            fontupper=\small,
            colback=PromptBack1,    
            colframe=PromptFrame1,  
            coltitle=white,      
        ]
            \textbf{Sysyem Prompt}
    
            You are an expert AI assistant specializing in multimodal long document understanding. Your task is to carefully analyze the provided memory of document information, and provide a precise answer to the user’s question.\\
            Follow these instructions carefully:
            \begin{itemize}[label= -,leftmargin=2em, nosep]
            \item Core Objective: Your primary goal is to accurately and concisely answer the user’s question based on the content of the memory and the reasoning chains.
            \item The memories are listed one by one. You need to understand each of them in sequence and do not omit any of them.
            \item If the information is sufficient, generate the final answer directly based on the memory.
            \item If the information is insufficient and briefly describe what information is missing.
            \item If you are certain that the document does not contain any information related to the user's question, please output Not answerable.
            \end{itemize}
            \medskip
            \textbf{User Prompt}
    
            Current Question:
            \{question\}
            
            Current Memory:
            \{memory\}

            Based on the above information, determine whether the memory have well-supported reasoning chains and does not require the retrieval of additional information. If sufficient, provide the answer directly; if insufficient, output according to the system prompt specification.\\
            Evaluate if current memory is sufficient to answer the question:\\
            1. Memory Completeness:
            \begin{itemize}[label= -,leftmargin=2em, nosep]
                \item What information has been successfully gathered?
                \item Are there obvious text sections that should have been examined?
                \item Does memory address core question requirements?
            \end{itemize}
            2. Information Gaps:
            \begin{itemize}[label= -,leftmargin=2em, nosep]
               \item What specific information is still missing?
               \item Which sections or data points need investigation?
               \item What additional searches would be beneficial?
            \end{itemize}
            \textbf{Important guidelines}
            \begin{itemize}[label= -,leftmargin=2em, nosep]
            \item If the memory does not mention the relevant topic at all: the answer is ``Not answerable''
            \item Rule of faithfulness: Be faithful. If the provided memory do not contain sufficient information to answer the user’s question, you should answer'Not answerable'.
            \end{itemize}
            \{Incontext few shot\}\\
            \textbf{Output Format}\\
            1. When information is sufficient: Output the final answer inside <final\_result></final\_result> tags. Do not add any extra text, prefixes, or explanations outside the tags.\\
            2. When information is insufficient: if additional clarification is needed, output the briefly describe the missing information, one item per line inside <missing\_info></missing\_info> tags.\\
            3. When the document does not contain any information related to the user's question:Output 'Not answerable' inside <final\_result></final\_result> tags. Do not add any extra text, prefixes, or explanations outside the tags.\\
        \end{tcolorbox}
        \caption{Prompt for \textbf{Reflector}.}
        \label{fig:Reflector_prompt}
    \end{figure*}

\end{document}